\documentclass[lettersize,journal]{IEEEtran}
\usepackage{amsmath,amsfonts}
\usepackage{algorithmic}
\usepackage{algorithm}
\usepackage{array}
\usepackage[caption=false,font=normalsize,labelfont=sf,textfont=sf]{subfig}
\usepackage{textcomp}
\usepackage{stfloats}
\usepackage{url}
\usepackage{verbatim}
\usepackage{graphicx}
\usepackage{cite}
\usepackage{colortbl}
\usepackage{tabularx}
\usepackage{multirow}
\usepackage{arydshln}
\usepackage{float}
\usepackage{booktabs}
\usepackage{graphicx}
\usepackage{hyperref}
\usepackage{amssymb}
\usepackage{mathrsfs}
\usepackage{amsmath}

\begin{document}

\title{Multi-Task Learning with Multi-Query Transformer for Dense Prediction}

\author{Yangyang Xu, Xiangtai Li, Haobo Yuan, Yibo Yang,
Lefei Zhang,~\IEEEmembership{Senior Member,~IEEE}
\thanks{Yangyang Xu, Haobo Yuan and Lefei Zhang are with the School of Computer Science, Wuhan University; E-mail: {yangyangxu@whu.edu.cn, yuanhaobo@whu.edu.cn, zhanglefei@whu.edu.cn.}}
\thanks{Xiangtai Li is with the Key Laboratory of Machine Perception, MOE, School of Artificial Intelligence, Peking University; E-mail: {lxtpku@pku.edu.cn.}}
\thanks{Yibo Yang with the JD Explore Academy; E-mail: {ibo@pku.edu.cn.}}
\thanks{Yangyang Xu and Xiangtai Li contribute equally.

Corresponding author: Lefei Zhang.}
\thanks{Manuscript received xx xx, 2023; revised xx xx, 2023.}}

\markboth{Journal of \LaTeX\ Class Files,~Vol.~14, No.~xx, xx~2023}%
{Shell \MakeLowercase{\textit{et al.}}: A Sample Article Using IEEEtran.cls for IEEE Journals}


\maketitle

\begin{abstract}
Previous multi-task dense prediction studies developed complex pipelines such as multi-modal distillations in multiple stages or searching for task relational contexts for each task.
The core insight beyond these methods is to maximize the mutual effects of each task. 
Inspired by the recent query-based Transformers, we propose a simple pipeline named Multi-Query Transformer (MQTransformer) that is equipped with multiple queries from different tasks to facilitate the reasoning among multiple tasks and simplify the cross-task interaction pipeline. 
Instead of modeling the dense per-pixel context among different tasks, we seek a task-specific proxy to perform cross-task reasoning via multiple queries where each query encodes the task-related context.
The MQTransformer is composed of three key components: shared encoder, cross-task query attention module and shared decoder. 
We first model each task with a task-relevant query. Then both the task-specific feature output by the feature extractor and the task-relevant query are fed into the shared encoder, thus encoding the task-relevant query from the task-specific feature. 
Secondly, we design a cross-task query attention module to reason the dependencies among multiple task-relevant queries; this enables the module to only focus on the query-level interaction.
Finally, we use a shared decoder to gradually refine the image features with the reasoned query features from different tasks.
Extensive experiment results on two dense prediction datasets (NYUD-v2 and PASCAL-Context) show that the proposed method is an effective approach and achieves state-of-the-art results. 
Code and models are available at \url{https://github.com/yangyangxu0/MQTransformer}.

\end{abstract}

\section{Introduction}
\label{sec:intro}

\noindent
\IEEEPARstart{H}{umans} are excellent at accomplishing multiple tasks simultaneously in the same scene. In computer vision, Multi-Task Dense Prediction (MTDP)~\cite{Multi-Modal_2021,multitask_mtst_2021} requires a model to directly output multiple task results such as semantic segmentation, depth estimation, normal prediction and boundary detection. High-performance MTDP results are important for several applications, including robot navigation and planning. Previous works use Convolution Neural Networks (CNNs) to capture different types of features for each task. Several approaches~\cite{multitask_UM_2019,Pad-net_2018,Mti-net_2020,mtlP_2021,atrc_2021} exploring task association achieved impressive results in MTDP.
Recently, transformer-based methods have achieved promising results on various vision tasks~\cite{ViT2021,InstanceSegmTransf_2021,swin}.
However, how to effectively learn and exploit the task-relevant information is still a promising direction.

\begin{figure*}[!t]
\centering
  \includegraphics[width=0.98 \textwidth]{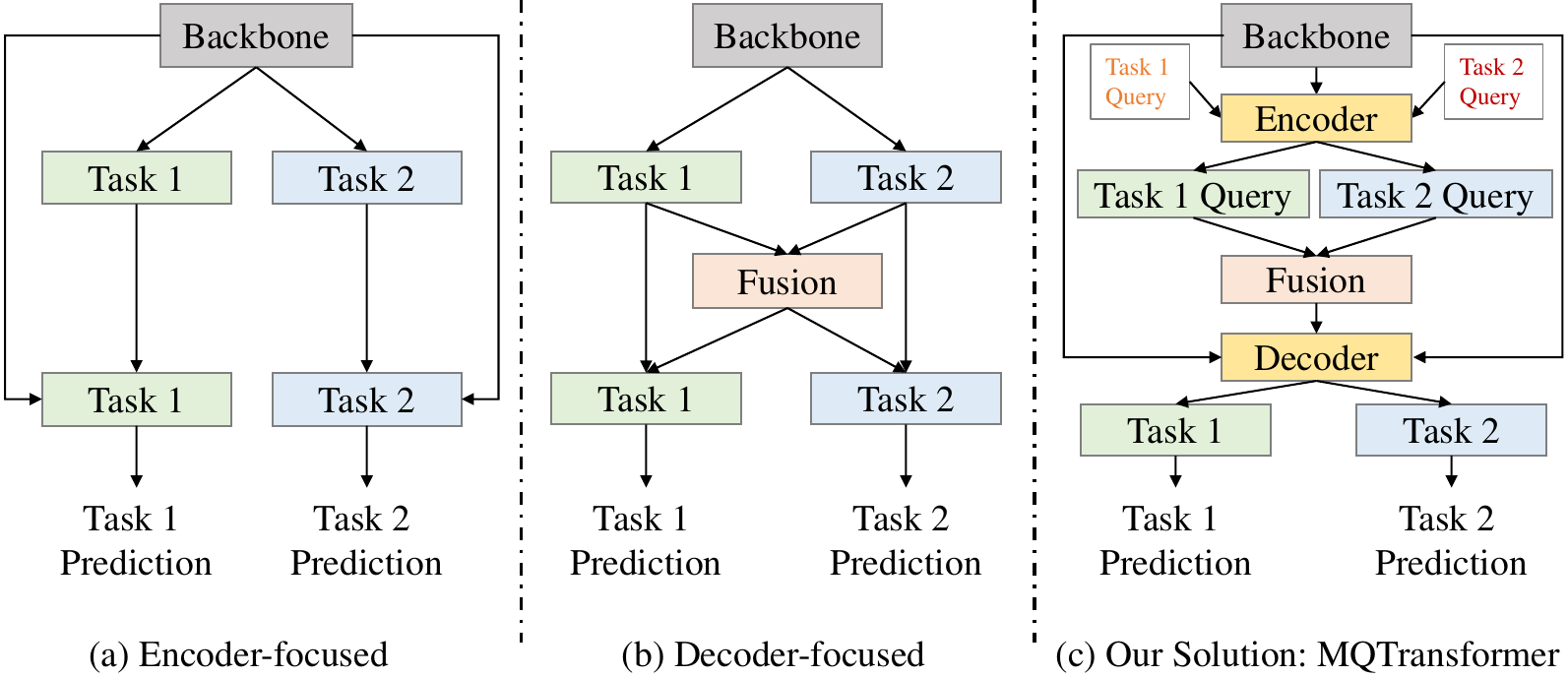}
  \caption{Illustration of different approaches for solving the MTDP task. We separate the encoder-focused and decoder-focused models based on where the task fusion occurs. (a) The baseline method proposed in~\cite{MTL_attn_2019} selects features from the shared encoder. (b) The fusion in (b) is performed in different manners, such as spatial attention~\cite{PAP-Net_2019}, distillation \cite{Pad-net_2018,atrc_2021}. (c) The proposed method adopts multiple task-relevant queries with Transformer and performs joint learning among different queries with a shared encoder and decoder.} 
  \label{fig1}
\end{figure*}

Current MTDP methods are accomplished by learning a shared representation for multi-task features and can be categorized into the \textit{encoder-focused} (Fig.~\ref{fig1}(a)) and \textit{decoder-focused} (Fig.~\ref{fig1}(b)) methods based on \textit{where the fusion of task-specific features occurs}. As shown in Fig.~\ref{fig1}(a), the encoder-focused methods~\cite{MTL_attn_2019,encoder_base_2018,densePredic_pvt_2021} share a generic feature and each task has a specific head to perform task prediction.
However, encoder-focused methods result in each task being individual and there are no task association operations. To this end, the decoder-focused models~\cite{atrc_2021,PAP-Net_2019,Pad-net_2018,Mti-net_2020} focus on the relationships among tasks via a variety of approaches. Neural architecture search (NAS)~\cite{atrc_2021,Adashare_nas_2020,yang2021towards} and knowledge distillation ~\cite{densePredic_cwkdis_2021} techniques are leveraged to find complementary information via the associated task sharing.
For example, the work~\cite{Adashare_nas_2020} is designed for determining effective feature sharing across tasks. Decoder-focused models for MDPT are usually of high computational cost due to the multiple states and roads needed for the interaction, such as the multi-modal distillation module of ATRC~\cite{atrc_2021}. Moreover, decoder-focused methods often contain complex pipelines and need specific human design. 

Recently, several transformer models~\cite{detr_2020} show simpler pipelines and promising results. In particular, the dense prediction Transformer (DPT)~\cite{Ranftl2021} exploits vision Transformers as a backbone for dense prediction tasks. The encoder-decoder architecture with object query~\cite{transformer2017,detr_2020} mechanism is proved to be effective in reasoning the relations among multiple objects and the global feature context. The object query design jointly tackles the interaction in two aspects: the relationship among queries and the interaction between feature and query. These successes inspire us to explore the potential of the multi-query Transformer with multiple task-relevant queries for multiple tasks learning where each query represents one specific task. Each task can be correlated via query reasoning among different tasks.
This query-based approach takes into account both the interaction between objects and making each query more explicitly associated with a specific spatial location and thus achieving an understanding of the full scene.

In this work, we introduce the Multi-Query Transformer (MQTransformer) for multi-task learning of six dense prediction tasks, including semantic segmentation, human parts segmentation, depth estimation, surface normal prediction, boundary detection, and saliency estimation.
There are three key designs.
Firstly, as illustrated in Fig.~\ref{fig1} (c) (setting two tasks for illustration purposes), we leverage multiple task-relevant queries (according to task number) as the input of the encoder. The encoder outputs the learnt task-relevant query feature of each task. 
Secondly, we design an efficient cross-task query attention module to facilitate the interaction of these task-relevant queries; this enables the cross-task query attention module to only focus on the query level, therefore, reducing complexity (see Tab.~\ref{Table:complexity}).
Thirdly, the task-relevant queries after cross-task query attention serves as the input of the following shared decoder. The shared decoder outputs the corresponding task-specific feature maps according to the number of tasks and applies them separately to task prediction. 

\begin{figure}[!t]
\centering
  \includegraphics[width=0.49 \textwidth]{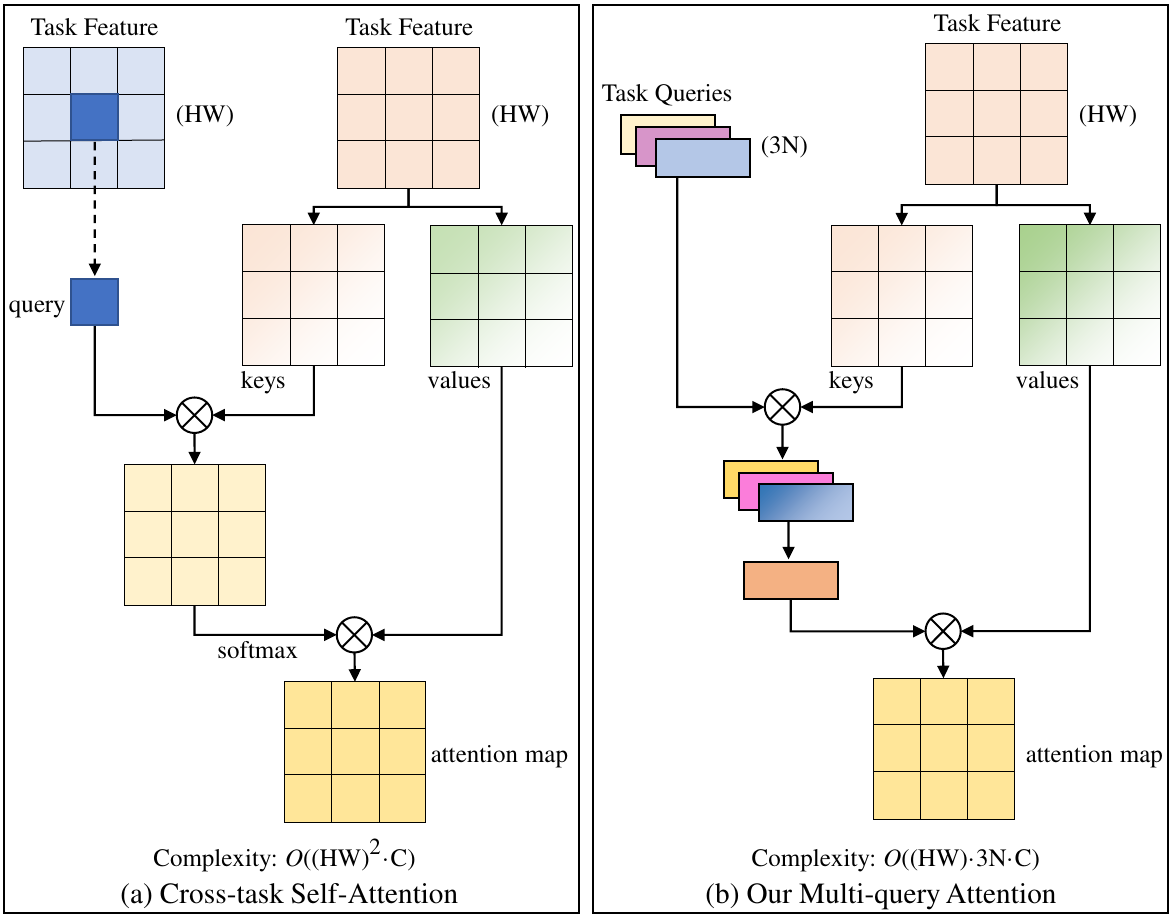}
  \caption{The structure of different attention mechanisms for MTDP: (a) The cross-task self-attention mechanism with pixel-wised affinity learning. 
  (b) Our model adopts multiple task-relevant queries with attention. Different colors represent different tasks. By replacing pixel-level affinity calculation across tasks, our method introduces task-relevant queries to encode task-aware context and perform cross-task query learning in an efficient manner.} 
  \label{fig2_our_attn}
\end{figure}

Since the task association is performed on the query level, we avoid heavy pixel-level context computation as used in previous work~\cite{atrc_2021}.
As demonstrated in Fig.~\ref{fig2_our_attn}, we show the difference between our multi-query attention (Right) and cross-task self-attention (Left) in more detail. Fig.~\ref{fig2_our_attn} (a) shows the query, key and value in self-attention from the different task features. Self-attention first calculates the dot product of the query with keys and then uses a softmax function to obtain the attention map on the value. We initialize independent multiple task-relevant queries as the attention mechanism input of the query and the key and value come from the task feature. We compute the task queries and key to obtain task queries with a valuable feature. Finally, we compute the dot product of the task queries and value, resulting in the final values, as depicted in Fig.~\ref{fig2_our_attn} (b). In addition, the computational complexity of our method is more friendly than cross-task self-attention. 

In our experiments, our method is compatible with a wide range of backbones, such as CNN~\cite{HRnet_19} and vision Transformer~\cite{swin,vitaev2}. We show the effectiveness of our query interaction among different tasks in various settings on different task metrics. Moreover, our experimental results demonstrate that the MQTransformer achieves better results than the previous methods in Fig.~\ref{fig1} (a) and (b). Our finding demonstrates that multiple task-relevant queries play an important role in MTDP. Another key insight is that the interaction of query features captures dependencies for different tasks. The main contributions of this work can be summarized as follows:

\begin{itemize}
\item [1)] We propose a new method, named MQTransformer, which effectively incorporates multiple task-relevant queries and task-specific features of the feature extraction for multi-task learning of dense prediction tasks, resulting in a simpler and stronger encoder-decoder network architecture.
\textit{To the best of our knowledge, this is the first exploration of multi-task dense prediction with a task-relevant query-based Transformer.}

\item [2)] We present the cross-task query attention module to enable sufficient interaction of the multiple task-relevant queries across tasks. Thus, the dependencies carried by each task query feature can be maximally refined. In addition, cross-task query attention only focuses on the query-level features, therefore, reducing complexity.

\item [3)] We conduct extensive experiments on two dense prediction datasets: NYUD-v2 and PASCAL-Context. Extensive experiment results show that MQTransformer consistently outperforms several competitive baseline methods. Notably, the proposed model exhibits \textit{significant} improvements for different tasks compared with the latest state-of-the-art results.
\end{itemize}

\section{Related Work}
\label{sec:relatedwork}

\noindent
\textbf{Multi-Task learning for Dense Prediction (MTDP).}
As deep neural networks have gradually become the mainstream framework for computer vision, multi-task learning has also developed tremendously.
Multi-task learning is typically used when related tasks can make predictions simultaneously. Many multi-task learning models~\cite{2010:JA,multitask_UM_2019,pattern_struct_diffusion_2020,eth_RC_2020,Cross-Stitch_2016,2019:SG,2018:ZZ,Mti-net_2020,multitask_mtst_2021}  have been widely used in various computer vision tasks. Recent work like \cite{pattern_struct_diffusion_2020} improves multi-task learning performance by co-propagating intra-task and inter-task pattern structures into task-level patterns, encapsulating them into end-to-end networks.
Furthermore, the work~\cite{atrc_2021} proposed an Adaptive Task-Relational Context (ATRC) module, which leverages a multi-modal distillation module to sample the features of all available contexts for each task pair, explicitly considering task relationships while using self-attention to enrich the features of the target task. 
The work~\cite{2021:GMI} designed an anomaly detection framework based on multi-task learning through self-supervised and model distillation tasks. 
CNN-based architectures~\cite{densePredic_dac_2018,densePredic_dcmd_2021,densePredic_fapn_2021} are proposed to capture the local information for the dense prediction task. In~\cite{densePredic_cwkdis_2021,atrc_2021}, the knowledge distillation is employed for dense prediction tasks.
NDDR-CNN~\cite{NDDR-CNN_2019} presents a CNN structure called neural discriminative dimensionality Reduction for multi-task learning, which enables automatic feature fusing at every layer and is formulated by combining existing CNN components in a new way with clear mathematical interpretability.
In MTI-Net~\cite{Mti-net_2020} network, the multi-scale multi-modal distillation unit is designed to perform task interaction on very scale features and the aggregation unit aggregates the task features from all scales to produce the final predictions for each task.
Recent work has attempted to use search learning to mold an optimal architecture.
The vast majority of these methods are trained using multi-scale features, where a grid search is typically used to select appropriate weights~\cite{atrc_2021}. 
Some works focus on building encoder-decoder networks to back-propagate high-level semantic contextual information from small-scale features to large-scale features through layer-by-layer up-sampling. 
Several works~\cite{yang2020sognet,li2020semantic} integrate the FPN~\cite{fpn2017} backbone bottom-up to generate multi-scale feature pyramids for dense prediction. 
Dense prediction transformer-based~\cite{densePredic_pvt_2021,Ranftl2021} encoder-decoder employs attention~\cite{transformer2017} computational operations to obtain fine-grained and globally consistent features to perform dense prediction tasks. In this paper, we explore Multi-Query Transformer for MTDP and propose a new method to adopt Vision Transformer into MTDP.

\noindent
\textbf{Vision Transformers.}
Currently, due to the unprecedented success of the Transformer~\cite{transformer2017} in natural language processing (NLP), many computer vision efforts are enthusiastically applying the Transformer to vision tasks~\cite{2021TransPose,2021TransDet,2021TransReS,2021TransTrac,2021TransPoint4D,2021TransClass}, starting with the Vision Transformer (ViT)~\cite{ViT2021}. The image is segmented into a fixed number of patches and they are embedded into a “token” as input. And project them into the feature space, where the converter encoder computes the queries, keys and values to generate the final result~\cite{attention_survey}. 
The encoder-decoder-based design Transformer has been applied to object detection, image classification and instance segmentation tasks~\cite{detr_2020,swin,InstanceSegmTransf_2021} and has demonstrated the great potential of attention-based models.
Various Transformer variants are based on self-attention and leveraged to serve the computer vision, such as Deformable DETR~\cite{deformableDetr}, T2T-ViT~\cite{2021:YL}, PVT~\cite{densePredic_pvt_2021}, Swin-Transformer~\cite{swin} and ViTAE~\cite{vitae}. 
DeiT~\cite{2021DeiT} further extended ViT by employing a new distillation method in which the Transformer learns more from images than others with similar Transformers. Moreover, several vision transformers~\cite{InstanceSegmTransf_2021,yuan2021polyphonicformer,video_knet,fashionformer,panopticpartformer,2022vsa} adopt DETR-like architecture to simplify the complex pipeline. Unit~\cite{hu2021unit} proposes to learn cross-modal and cross-task using the DETR-like model.
Concerning dense scene understanding tasks, attention mechanisms approach to efficiently maintain multi-scale features in the network.
MTFormer~\cite{xu2022mtformer} designs the shared transformer encoder and shared transformer decoder, and task-specific self-supervised cross-task branches are introduced to obtain final outputs, which increases the MTL performance. 
InvPT~\cite{InvPT_2022} is designed to learn the long-range interaction in both spatial and all-task contexts on the multi-task feature maps with a gradually increased spatial resolution for multi-task of dense prediction.
Recently, these successful works are designed using Transformer architecture to learn good representations for multiple vision tasks.

\noindent
Differing from previous models~\cite{Ranftl2021,InvPT_2022,xu2022mtformer}, our work explores vision Transformers for multi-task representation and is complementary to these efforts where all the vision Transformers serve as the feature extractor or instance-level learner. 
Moreover, different from the object query in DETR~\cite{detr_2020}, which represents the object in the scene, our task-relevant queries explore the relationship context among different tasks where the task-specific features are already encoded before reasoning.

\section{Method}
\label{sec:method}

\begin{figure*}[ht]
\centering
  \includegraphics[width=0.95\textwidth]{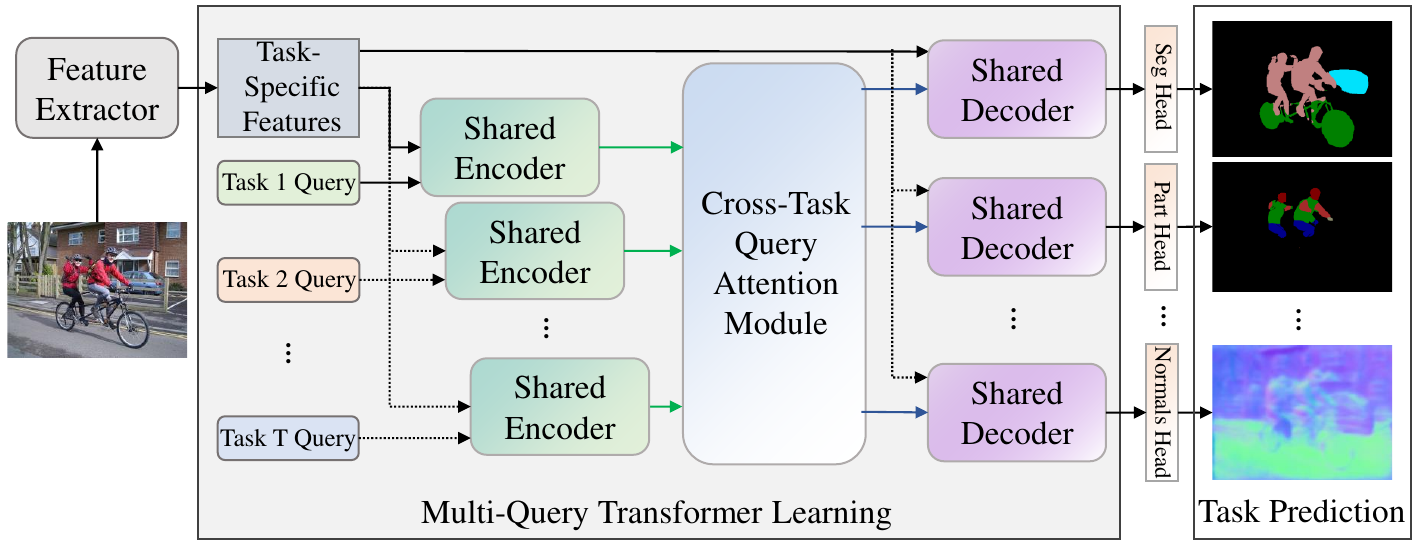}
  \caption{\small An overview of MQTransformer. The MQTransformer represents multiple task-relevant queries to extract task-specific features from different tasks and performs joint multi-task learning.
  Here, we show an example of task-specific policy learned using our method. Note that the encoder (aqua) and decoder (mauve) are shared in our model.
  The number of task queries depends on the number of tasks. There are $T$ tasks.
  The task query first generates uniform initialization weights and then applies these task queries to encode from the corresponding task-specific feature in the shared encoder.
  The 'Seg', 'Part' and 'Normals' mean semantic segmentation and human part segmentation and surface normals tasks, respectively.
  }
  \label{fig:over_view}
\end{figure*}

\noindent
\textbf{Overview.} In this part, we will first introduce the problem formulation and motivation of our approach in Sec.~\ref{sec:problem_formulation}. We present the details of our method and insights in Sec.~\ref{sec:main_method}.
Then we give the description of the loss function of our architecture in Sec.~\ref{sec:loss}.

\subsection{Problem Formulation and Motivation}
\label{sec:problem_formulation}

To facilitate the description of the components in the model, we first briefly introduce the basic notation of the Transformer~\cite{transformer2017}. For short, we term these operations including Multi-Head Self-Attention (MHSA), Multi-Layer Perceptron (MLP) and Layer Normalisation (LN). 
An image $x \in \mathbb{R}^{H \times W \times 3}$ is fed into the feature extractor and then generates a task-specific feature $X \in \mathbb{R}^{H/4 \times W/4 \times C}$.
There are two kinds of features described in our model.
One is the task-specific feature $X \in \mathbb{R}^{H/4 \times W/4 \times C}$ and the other is the task query $p \in \mathbb{R}^{N \times C}$.
$H$, $W$ and $C$ are the height, width and channel.
$N$ is a pre-defined constant number.
We perform reshape operation to flatten the task-specific feature $ \mathbb{R}^{H/4 \times W/4 \times C}$ to a 1D sequence $\mathbb{R}^{L \times C}$ ($L=H/4 \times W/4$). 
$L$ is the number of pixels of a task-specific feature.

The goal of MTDP is to directly output several independent dense prediction maps. Previous works, including decoder-focused and encoder-focused, pay more attention to the design space of a cross-task query interaction.
There are two main problems: one is the huge computation cost and affinity memory in the case of the cross-task, even though only one scale is considered. The other is the complex pipeline for modeling each decoder, such as NAS. Moreover, \textit{how to adapt a vision transformer for MTDP is still an open question.} To tackle those two problems and explore a new vision transformer architecture for MTDP, we present a multi-query Transformer.

\subsection{Multi-Query Transformer (MQTransformer)}
\label{sec:main_method}

\noindent
\textbf{Overview.} As shown in Fig.~\ref{fig:over_view}, our MQTransformer contains feature extractor and multi-query transformer learning. The former extracts features, while the latter is a multi-query transformer to perform cross-task query association. The latter contains a shared encoder, a cross-task query attention module and a shared decoder where the task-relevant queries are the inputs and perform that task association. 

\noindent
\textbf{Feature Extractor.} We first extract image features for each input image. It contains a backbone network (Convolution Neural Network~\cite{HRnet_19} or Vision Transformer~\cite{swin}). This results in a set of multi-scale features. We fuse these features into one single high-resolution feature map (task-specific feature in Fig.~\ref{fig:over_view}) via addition and bilinear interpolation upsampling.

\begin{figure*}[!ht]
\centering
  \includegraphics[width=0.95\textwidth]{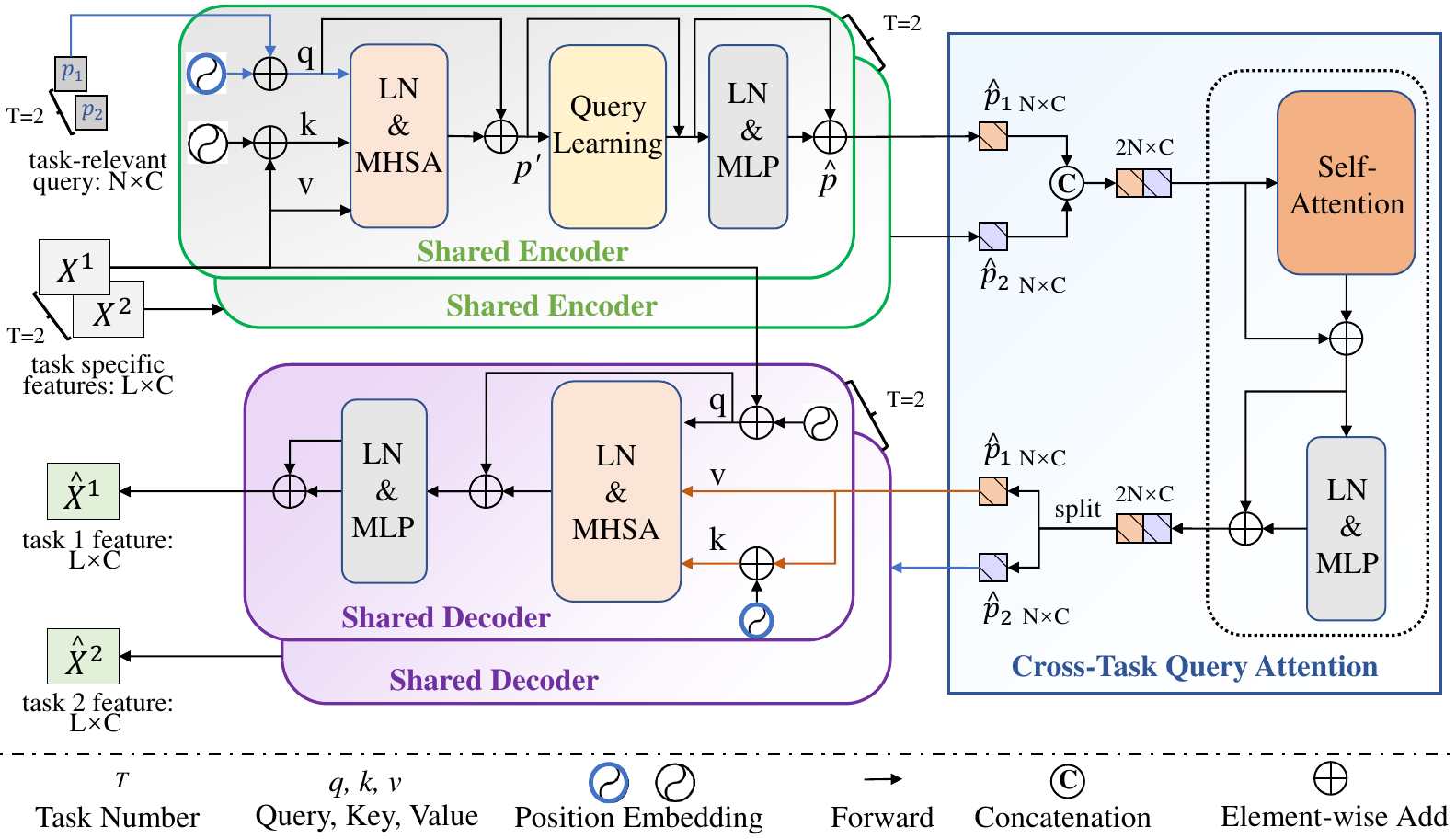}
  \caption{\small Illustration of the Multi-Query Transformer (MQTransformer). We use a shared encoder to obtain the valuable task-relevant query through the input of task-specific features $X^t$ and task-relevant query $p_{t}$ ($t$ denotes the task number). The task-relevant queries $\hat p$ after the upstream operation are concatenated by cross-task. The cross-task query attention module leads to the concatenated task-relevant query interaction via self-attention. Then, the task-relevant query is split into a shared decoder where it outputs the task-specific features $\hat X^t$ for final predictions. To simplify these pipelines, we demonstrate two tasks ($i,e., T=2$).
}
  \label{fig_encoder_decoder}
\end{figure*}

\noindent
\textbf{Motivation of Multi-Query Transformer.} Our key insight is to replace complex pixel-wised task associations with task-relevant queries. To achieve that, we use a shared encoder to encode each task-specific feature into their corresponding queries. Then the task association can be performed within these queries. Finally, the refined task-specific features can be decoded from the reasoned queries with another shared decoder.

\noindent
\textbf{Shared Encoder.} 
We feed the extracted task-specific feature and the task-relevant query into a shared encoder. Such encoder builds the one-to-one connection between features and queries on different tasks, which is shown in the pink box in Fig.~\ref{fig_encoder_decoder}. 

\noindent
The task-relevant query $p\in \mathbb{R}^{N \times C}$ vector is uniformly initialized and is not further optimized.
A task-relevant query corresponds to just one task. 
Normally we will produce the number of task-relevant queries according to the number of tasks.
The learned positional encodings~\cite{transformer2017} $e_q\in \mathbb{R}^{N \times C}$ (see Fig.~\ref{fig_encoder_decoder} blue circle) are added to the task-relevant query before regularization using LN. 
Then they are used as a \textit{query} input for the MHSA in the shared encoder.
The task-specific feature $X \in  \mathbb{R}^{H/4 \times W/4 \times C}$ from the feature extractor is reshaped to $x \in \mathbb{R}^{L \times C}$, and added to positional encoding $e_k\in \mathbb{R}^{L \times C}$ (see Fig.~\ref{fig_encoder_decoder} black circle), and then input to the MHSA as \textit{key} and \textit{value}.
The formulation of MHSA is
   \begin{align}\label{1}
    	&\text{MHSA}(Q, K, V) = \text{softmax}(QK^{T}/ \sqrt{d})V,
    \end{align}
\noindent 
where $Q \in \mathbb{R}^{N \times C}$, $K \in  \mathbb{R}^{L \times C}$ and $ V \in  \mathbb{R}^{L\times C}$ are the query, key and value tensors; $d$ denotes dimension $C$; $\text{MHSA}(Q, K, V) \in \mathbb{R}^{N \times C}$.
And then, the encoder is calculated as follows, where task-specific features are encoded into query format:
\begin{align}\label{3}
p' = p  + \text{MHSA}(Q=\text{LN}(p), K=\text{LN}(X), V=\text{LN}(X)).
\end{align}

\begin{table*}[ht]
\centering
  \caption{
    Complexity comparison. We compare the computational complexity of the different schemes for cross-task communication. $H$ and $W$ are the image height and width. $C$, $N$, and $K$ are hyper-parameters. We adopt $C=256$, $N=100$, and $K=9$ when calculating GFLOPs.
    Our cross-task query attention is not perceptive to the resolution of the image.
    }
  \label{Table:complexity}
  \setlength{\tabcolsep}{0.9mm}{
  \centering
  \begin{tabular}{lccc}
    \toprule
    \multirow{2}{*}{Method}     & \multirow{2}{*}{Complexity}         & \multicolumn{2}{c}{GFLOPs}\\
    & & $64 \times 64$ & $128 \times 128$ \\
    \midrule
    No Communication                         & 0         & 0      & 0    \\
    Global Context~\cite{atrc_2021,ViT2021} & $\mathcal{O}((HW)^2C)$  &    9.74   &   142.83   \\
    Local Context~\cite{atrc_2021,swin}  &    $\mathcal{O}(HWC^2 \cdot K^2)$ & 21.74      & 86.98       \\
    Cross-Task Query Attention (Ours)    &    $\mathcal{O}(C \cdot N^2)$ & 0.03      & 0.03    \\
    \bottomrule
  \end{tabular}}
\end{table*}

\noindent
\textbf{Query Learning in Shared Encoder.} Then the task-relevant query $\hat p$ captures the task-relevant context. We use a linear layer (linear mapping) along with an identity connection as:
  \begin{align}\label{5}
     p' = p' + \text{Linear}(\text{LN}(p')),
   \end{align}

\noindent
where the $\hat p \in \mathbb{R}^{N \times C}$ is the result after the query learning step. To enhance query learning, we adopt an extra layer norm and MLP (a non-linear GELU between the two linear layers) with an identity connection to update the task-relevant query. The projection can be formulated as follows:

  \begin{align}\label{eq:MLP}
    \hat{p} =  p' + \text{MLP}(\text{LN}({p'})).
   \end{align}

This operation can further reduce the inductive biases and enhance the communications in query. Each task-relevant query is processed independently and only gradually interacts with the image feature to learn dependencies. 

\noindent
As shown in Fig~\ref{fig_encoder_decoder} cross-task query attention module, we take task-relevant queries with two tasks for illustration. Each task-relevant query follows the same pipeline above. In the end, we obtain two different queries ($\hat p_{1}$ and $\hat p_{2}$). 

\noindent 
\textbf{Cross-Task Query Attention Module.} Cross-Task Query Attention module aims to perform cross-task query learning for task-relevant queries ($\hat p_{1}$ and $\hat p_{2}$). 
We concatenate the task-relevant queries of the same from all tasks: $P_{cat}=\text{Concat}(\hat p_{1}, \hat p_{2})$.
$P_{cat} \in \mathbb{R}^{(N+N) \times C}$ is a cross-task query.
Then we perform MHSA operation as follows:

\begin{align}\label{6}
P_{cat}' &= P_{cat}  + \text{MHSA}(Q=P_{cat}, K=P_{cat}, V=P_{cat}), \\
\hat P_{cat} &= P_{cat}'  + \text{MLP}(\text{LN}(P_{cat}')).
\end{align}

The cross-task query attention module is designed to enhance the communication of multiple task-relevant queries.
After the task-relevant query interaction for different tasks, we split them according to the number of tasks for downstream operation.
We split the result $ \hat P_{cat} \in \mathbb{R}^{(N+N) \times C}$ into $\hat p_{1} \in \mathbb{R}^{N \times C}$ and $ \hat p_{2} \in \mathbb{R}^{N \times C}$.

Our cross-task query attention module builds the relation from different tasks via query-level association. Compared with previous pixel-level or patch-level cross-task learning, it avoids heavy affinity costs. Tab.~\ref{Table:complexity} shows the complexity and GFLOPs of query-based attention. Our task-relevant query interaction reduces the required computation (Our cross-task query attention: 0.03 GFLOPs v.s. ATRC~\cite{atrc_2021}(global context): 9.74 GFLOPs) with the $64\times64$ image feature as inputs. 
Because our cross-task query attention module complexity can only perceive $N$ and $C$ dimensions but not the resolution of the image.

\noindent
\textbf{Shared Decoder.} 
We propose a shared decoder to model the relations among \textit{queries, keys and values} effectively under the guidance of the task-relevant query in the task-specific feature.
There are two inputs for the shared decoder for each task, the learned task-relevant query $ \hat p_{1}$ and the task-specific feature $X$.
As shown in Fig.~\ref{fig_encoder_decoder} green box, the shared decoder contains a MHSA, LN and MLP.
MHSA is applied to interact between the learned task-relevant query and the task-specific feature.  
We use the learned task-relevant query $\hat p$ as \textit{key/value} and the task-specific feature $X$ as \textit{query} in MHSA. 
We leverage the $\hat p$ and $X$ to perform cross-attention~\cite{transformer2017}.
In particular, the $X$ (\textit{query}) and $\hat{p}$ (\textit{key/value}) are input into the MHSA.
Therefore, each pixel of the task-specific feature $X$ can have a long-range interaction with the learned task-relevant query.
The learned task-relevant queries and image feature interactions in the shared decoder (Fig.~\ref{fig_encoder_decoder} green box) can be written as follows: 
\begin{align}\label{8}
X' &= X  + \text{MHSA}(Q=\text{LN}(X), K=\text{LN}( \hat p), V=\text{LN}(\hat p)),\\
\hat X &= X' + \text{MLP}(\text{LN}(X')),
\end{align}
where $\hat X \in \mathbb{R}^{{L} \times C}$ is the task-specific feature.
Then we conduct a reshape operation to reshape the sequence $\hat X \in \mathbb{R}^{{L} \times C}$ back to task-specific feature $\hat X \in \mathbb{R}^{H/4 \times W/4 \times C}$.
Note that the encoder is shared across different tasks.

However, we argue that task-relevant queries can absorb relevant features in a compact manner without the extra parameters brought by encoding or decoding for this purpose.
We verify this in Sec.~\ref{sec:exp}.

\subsection{Loss Function}
\label{sec:loss}

After adopting prediction heads for each task via $1\times1$ convolution, we employ task-specific loss functions for each task. For semantic segmentation, human part segmentation and saliency detection, the cross-entropy loss is adopted. We use L1-Loss and balance binary cross-entropy losses for the depth, surface normals and boundary supervision, respectively. Then the final training objective ${\mathcal L}$ for each task can be formulated as follows:
	\begin{equation}\label{equ:loss}
	\begin{split}
	    \begin{aligned}
    {\mathcal L} &= \lambda_{seg} {\mathcal L_{seg}} + \lambda_{depth} {\mathcal L_{depth}} + \lambda_{normals} {\mathcal L_{normals}}\\ 
    &+ \lambda_{bound} {\mathcal L_{bound}} + \lambda_{partseg} {\mathcal L_{partseg}} + \lambda_{sal} {\mathcal L_{sal}},\\
       \end{aligned}
    \end{split}
	\end{equation}
\noindent 
where ${\mathcal L}$ denotes a total loss function. We set $\lambda_{seg}$=1.0, $\lambda_{depth}$=1.0, $\lambda_{normals}$= 10.0, $\lambda_{bound}$=50.0, $\lambda_{partseg}$=2.0, and $\lambda_{sal}$=5.0. For fair comparison, we adopt the same setting as ATRC~\cite{atrc_2021}.

\section{Experiment}
\label{sec:exp}

\subsection{Setup}
\noindent
\textbf{Datasets.} Following previous works~\cite{atrc_2021} in MTDP, we use NYUD-v2~\cite{NYUD2012} and PASCAL-Context~\cite{pascal2014} datasets.
The NYUD-v2 dataset is comprised of video sequences of 795 training and 654 testing images of indoor scenes.
It contains four tasks, including semantic segmentation (SemSeg), Monocular depth estimation (Depth), surface normals prediction (Normals) and semantic boundary detection (Bound). PASCAL-Context contains 4,998 training images and 5,105 testing images. 
It contains five tasks, including semantic segmentation (SemSeg), human part segmentation (PartSeg), saliency detection (Sal), surface normals prediction (Normals), and semantic boundary detection (Bound).

\begin{table*}[!tp]
  \centering
\caption{Comparison results on NYUD-v2 dataset. Notation ‘$\downarrow$’: lower is better. Notation ‘$\uparrow$’: higher is better.
"Params" denotes parameters. We report comparisons on various baseline models (in the first and second sub-figures) and several recent works. Single task baseline indicates a model corresponding to only a single task.
Swin-$\diamond$ indicates that the specific Swin model is uncertain. $\blacklozenge$ denotes results not reported in InvPT~\cite{InvPT_2022} but from our test. 
}
\label{tab:stoa_nyud_v2}
\begin{tabular}{llllllllr}
\hline\noalign{\smallskip}
 \multirow{2}*{Model} &\multirow{2}*{Backbone} &Params &GFLOPs &{SemSeg} &Depth &Normals &Bound &\multirow{2}*{$\Delta_m[\%]$$\uparrow$}\\
      &  &(M) &(G) &(mIoU)$\uparrow$   &(rmse)$\downarrow$   & (mErr)$\downarrow$  &(odsF)$\uparrow$\\
\noalign{\smallskip}
\hline
\noalign{\smallskip}

single task baseline &HRNet18   &16.09  &40.93   &38.02  &0.6104 &20.94 &76.22 &0.00\\
multi-task baseline  &HRNet18   &4.52   &17.59   &36.35  &0.6284 &21.02 &76.36 &-1.89\\
single task baseline &ViTAEv2-S &86.19  &152.46  &47.02  &0.5865 &20.98 &76.40 &0.00\\
multi-task baseline  &ViTAEv2-S &22.17  &76.44   &43.65  &0.5971 &21.02 &76.20 &-2.33\\
single task baseline &Swin-T    &115.08 &161.25  &42.92  &0.6104 &20.94 &76.22 &0.00\\
multi-task baseline  &Swin-T    &32.50  &96.29   &38.78  &0.6312 &21.05 &75.60 &-3.74\\
single task baseline &Swin-S    &200.33 &242.63  &48.92  &0.5804 &20.94 &77.20 &0.00\\
multi-task baseline  &Swin-S    &53.82  &116.63  &47.90  &0.6053 &21.17 &76.90 &-1.96\\

\hdashline
Cross-Stitch\cite{Cross-Stitch_2016}   &HRNet18 &4.52  &17.59 &36.34 &0.6290 &20.88 &76.38 &-1.75\\
Pad-Net\cite{Pad-net_2018}             &HRNet18 &5.02  &25.18 &36.70 &0.6264 &20.85 &76.50 &-1.33\\
PAP\cite{PAP-Net_2019}                 &HRNet18 &4.54  &53.04 &36.72 &0.6178 &20.82 &76.42 &-0.95\\
PSD\cite{pattern_struct_diffusion_2020}&HRNet18 &4.71  &21.10 &36.69 &0.6246 &20.87 &76.42 &-1.30\\
NDDR-CNN\cite{NDDR-CNN_2019}           &HRNet18 &4.59  &18.68 &36.72 &0.6288 &20.89 &76.32 &-1.51\\
MTI-Net\cite{Mti-net_2020}             &HRNet18 &12.56 &19.14 &36.61 &0.6270 &20.85 &76.38 &-1.44\\
ATRC\cite{atrc_2021}                   &HRNet18 &5.06  &25.76 &38.90 &0.6010 &20.48 &76.34 &1.56\\

\hdashline
MQTransformer (ours) & HRNet18  &5.23  &22.97  &40.47  &0.5965  &20.34  &76.60  &3.01\\
MQTransformer (ours) &ViTAEv2-S &26.18 &94.77  &48.37  &0.5769  &20.73  &76.90  &1.82\\
MQTransformer (ours) & Swin-T   &35.35 &106.02 &43.61  &0.5979  &20.05  &76.20  &0.31\\
MQTransformer (ours)         &Swin-S   &56.67 &126.37 &49.18  &0.5785  &20.81  &77.00  &1.59\\
MTFormer\cite{xu2022mtformer}&Swin-$\diamond$&64.03&117.73&50.56  &0.4830   &-      &-      &4.12\\
InvPT\cite{InvPT_2022}       &Swin-L &-      &-      &51.76  &0.5020  &19.39  &77.60  &-2.22$^{\blacklozenge}$\\
InvPT\cite{InvPT_2022} &ViT-L &402.1$^{\blacklozenge}$ &555.57$^{\blacklozenge}$ &53.56  &0.5183  &19.04  &78.10  &-\\
MQTransformer (ours)         &Swin-L &204.3      &365.25       &54.84   &0.5325  &19.67  &78.20  &-2.12\\ 
\hline
  \end{tabular}
\end{table*}

\begin{table*}[!tp]
  \centering
\caption{Results on the PASCAL-Context dataset. We also report comparisons on various baseline models (in the first and second sub-figure) and several recent works. The notation ‘$\downarrow$’: lower is better. The notation ‘$\uparrow$’: higher is better.}
\label{tab:stoa_pascal}
\begin{tabular}{lllllllr}
\hline\noalign{\smallskip}
 \multirow{2}*{Model} &\multirow{2}*{Backbone} &{SemSeg} &PartSeg &Sal &Normals &Bound &\multirow{2}*{$\Delta_m[\%]$$\uparrow$}\\
     &  & (mIoU)$\uparrow$  & (mIoU)$\uparrow$  &(maxF)$\uparrow$  & (mErr)$\downarrow$  & (odsF)$\uparrow$\\
\noalign{\smallskip}
\hline
\noalign{\smallskip}
Single task baseline  &HRNet18   &62.23  &61.66 &85.08  &13.69 &73.06 &0.00\\
multi-task baseline   &HRNet18   &51.48  &57.23 &83.43  &14.10 &69.76 &-6.77\\
single task baseline  &ViTAEv2-S &68.81	 &58.05 &82.89  &13.99 &71.10 &0.00\\
multi-task baseline   &ViTAEv2-S &64.66	 &56.65 &81.02  &14.93 &70.70 &-3.59\\
single task baseline  &Swin-T    &67.81	 &56.32 &82.18  &14.81 &70.90 &0.00\\
multi-task baseline   &Swin-T    &64.74	 &53.25 &76.88  &15.86 &69.00 &-3.23\\
single task baseline  &Swin-S    &70.83	 &59.71 &82.64  &15.13 &71.20 &0.00\\
multi-task baseline   &Swin-S    &68.10	 &56.20 &80.64  &16.09 &70.20 &-3.97\\
\hdashline
MTI-Net~\cite{Mti-net_2020} &HRNet18  &61.70 &60.18 &84.78  &14.23 &70.80 &-2.12\\
PAD-Net~\cite{Pad-net_2018} &HRNet18  &53.60 &59.60 &65.80  &15.3  &72.50 &-4.41\\
ATRC~\cite{atrc_2021} &HRNet18        &57.89 &57.33 &83.77  &13.99 &69.74 &-4.45\\ 
ASPP~\cite{ASPP_2018} &ResNet50       &62.70 &59.98 &83.81  &14.34 &71.28 &1.77\\
BMTAS~\cite{BMTAS_2020} &ResNet50     &56.37 &62.54 &79.91  &14.60 &72.83 &-0.55\\
DYMU~\cite{DYMU2022} &MobileNetV2     &63.60 &59.41 &64.94  &- &-         &0.18\\ 
ATRC~\cite{atrc_2021} &ResNet50       &62.99 &59.79 &82.25  &14.67 &71.20 &0.95\\

\hdashline
MQTransformer (ours) &HRNet18    &58.91 &57.43  &83.78 &14.17  &69.80 &-4.20\\
MQTransformer (ours) &ViTAEv2-S  &69.10	&58.23	&83.51 &13.73  &71.30 &0.72\\
MQTransformer (ours) &Swin-T     &68.24	&57.05	&83.40 &14.56  &71.10 &1.07\\
MQTransformer (ours) &Swin-S     &71.25	&60.11	&84.05 &14.74  &71.80 &1.27\\
\hline
  \end{tabular}
\end{table*}

\noindent
\textbf{Implementation details.}
Swin Transformer~\cite{swin} (Swin-T, Swin-S and Swin-B indicate Swin Transformer with tiny, small and base, respectively.), ViT-L~\cite{ViT2021} (ViT-L means ViT-Large), ViTAEv2-S~\cite{vitaev2} and HRNet (HRNet18, HRNet48)~\cite{HRnet_19} models are adopted as the feature extractor. 
We mainly perform ablation studies using Swin Transformer. Our model follows the initialization scheme proposed in ~\cite{atrc_2021}. We use Pytorch Framework to implement all the experiments in one codebase. During training, we augment input images during training by random scaling with values between 0.5 and 2.0 and random cropping to the input size. We deploy the SGD optimizer with default hyper parameters. The learning rate is set to 0.001 and weight decay is set to 0.0005. All the models are trained for 40k iteration steps with an 8 batch size on the NYUD-v2 dataset. For the PASCAL-Context dataset, we follow the same setting in NYUD-v2. All the models adopt a single inference for both ablation and comparison. 
Note that in practice, we use two-scale task-specific features (in Fig.~\ref{fig_encoder_decoder}) and use pyramid features ablation shown in Tab.~\ref{tab:ablation_scales}.

\noindent
\textbf{Strong Multi-task baselines.} The baseline model has the same architecture as MQTransformer. However, it does not contain MQTransformer prediction heads. It contains $T$ different heads for different tasks. For Swin Transformer, we adopt the ADE-20k pre-trained model as the strong baseline. For HRNet, we follow the same setting as ATRC~\cite{atrc_2021}. We argue that our baseline models are different from the previous method using Swin Transformer. However, even on such a strong baseline, our method can still achieve significant improvements. Moreover, we also prove the effectiveness and generation of our method on the CNN backbone in the ablation study (Sec.~\ref{sec:exp_ablation}).

\noindent
\textbf{Evaluation Metrics.}
Our evaluation follows the metrics scheme proposed in~\cite{atrc_2021}.
Semseg and PartSeg tasks are evaluated with the mean intersection over union (mIoU) and Depth task using the root mean square error (rmse). Normals task using mean error (mErr), Sal task using maximum F-measure (maxF), and Bound task using the optimal-dataset-scale F-measure (odsF). The average per-task performance drop ($\Delta_m$) is adopted to quantify multi-task performance. $\Delta_m=\frac{1}{T} \sum_{i=1}^T(M_{m,i}-M_{s,i})/M_{s,i}$, where m and s mean multi-task model and single task baseline. T denotes the total tasks. 
$\Delta_m$: the higher is the better.
GFLOPs and Parameters are often used to evaluate model efficiency.
To be specific, GFLOPs means floating point operations per second (the lower the better) and Parameters are used as an indirect indicator of computational complexity as well as memory usage (the lower the better).

\begin{table*}[!t]
\begin{center}
\linespread{0.8}
\scriptsize
	\centering
	\caption{Ablation studies and analysis on NYUD-v2 dataset using a Swin-S backbone. Query Learning (QL) and Cross-Task Query Attention (CTQA) Module are part of our model. Q\&C indicates QL and CTQA. HR48 denotes HRNet48. The notation ‘$\downarrow$’: lower is better. The notation ‘$\uparrow$’: higher is better. The \textbf{w/o} indicates \textbf{"without"}.} \label{tab:multi-ablaton}
	\setlength{\tabcolsep}{3.pt}
	\subfloat[\small Ablation on different modules]{
        \label{tab:module}
        \resizebox{0.450\textwidth}{!}{
		\begin{tabularx}{5.5cm}{l|c|c|c|c} 
		       \toprule
		       \multirow{2}*{Model}    & {SemSeg}   &Depth & Normals & Bound\\
                 & (mIoU)$\uparrow$   &(rmse)$\downarrow$   & (mErr)$\downarrow$  &(odsF)$\uparrow$\\
    	 	\toprule
         w/o QL    &48.97  &0.5807  &20.82  &76.1\\
         w/o CTQA  &48.93  &0.5824  &20.87  &75.6\\
         w/o Q\&C  &48.64  &0.5854  &20.87  &75.7\\
         Ours      &49.18  & 0.5785 &20.80  &77.0\\
			\bottomrule
		\end{tabularx} }
    } \hfill
    \setlength{\tabcolsep}{6.pt}
    \subfloat[\small Ablation on the depths ($D$) of our MQTransformer]{
        \label{tab:depth}
        \resizebox{0.460\textwidth}{!}{
		\begin{tabularx}{5.5cm}{c|c|c|c|c} 
		        				\toprule
    		    \multirow{2}*{$D$} & {SemSeg}   &Depth & Normals & Bound\\ 
    		   & (mIoU)$\uparrow$   &(rmse)$\downarrow$   & (mErr)$\downarrow$  &(odsF)$\uparrow$\\
    		 
    	 	\toprule
    	 
         1  &49.18    &0.5785   &20.80    &77.0\\
         2  &47.80    &0.6006   &21.08    &76.5\\
         4  &47.88	  &0.5983   &21.21    &76.5\\
			\bottomrule
		\end{tabularx}}
    } \hfill
    \setlength{\tabcolsep}{5.pt}
    \subfloat[\small Ablation on $N$. Query: $P\in \mathbb{R}^{N \times C}$]{
        \label{tab:part_N}
        \resizebox{0.450\textwidth}{!}{
		\begin{tabularx}{5.5cm}{c|c|c|c|c} 
		    \toprule
    		  \multirow{2}*{$N$} & {SemSeg}   &Depth & Normals & Bound\\ 
    		  	   & (mIoU)$\uparrow$   &(rmse)$\downarrow$   & (mErr)$\downarrow$  &(odsF)$\uparrow$\\

    	 	\toprule
    	 	8   &48.32  &0.5974  &20.90 &76.9\\
            32  &49.11	&0.5803  &20.58	 &77.0  \\
            64   &49.18  &0.5785  &20.80  &77.0 \\
            128  &49.63	 &0.5820  &20.84 &77.1  \\
            156  &48.81  &0.5941  &20.87 &76.8  \\
            256  &48.41	&0.6014   &21.01 &76.7  \\
			\bottomrule
		\end{tabularx}}
    } \hfill
    \subfloat[\small Ablation on backbones. MT-B means Multi-task Baseline]{
        \setlength{\tabcolsep}{2.pt}
        \label{tab:backbone}
        \resizebox{0.460\textwidth}{!}{
	    \begin{tabularx}{5.8cm}{l|l|c|c|c}
		 \toprule
    		 \multirow{2}*{Backbone}  & {SemSeg}   &Depth & Normals & Bound\\ 
    		 & (mIoU)$\uparrow$   &(rmse)$\downarrow$   & (mErr)$\downarrow$  &(odsF)$\uparrow$\\
    		\toprule
    	       HR48 (MT-B)                  &41.96 &0.5543 &20.36 &77.6 \\ 
    	       HR48+ATRC\cite{atrc_2021}    &46.27 &0.5495 &20.20 &77.6 \\
    	       HR48+Ours                    &47.48 &0.5374 &20.06 &78.0 \\
    	       \hdashline
    	       Swin-B (MT-B)                 &51.44   &0.5813  &20.44   &77.0 \\
                  Swin-B+InvPT~\cite{InvPT_2022}&50.97   &0.5071  &19.39   &77.3\\
    	       Swin-B+Ours                   &52.06   &0.5439  &19.85   &77.8 \\
        	\bottomrule
	    \end{tabularx}}
    } \hfill
\end{center}
\end{table*}

\subsection{Comparison with the state-of-the-art methods}
\noindent
\textbf{Results on NYUD-v2 dataset.} As shown in Tab.~\ref{tab:stoa_nyud_v2}, we report our MQTransformer results compared with both previous work~\cite{atrc_2021,Mti-net_2020} and strong multi-task baseline. Note that the transformer-based multi-task baselines in Tab.~\ref{tab:stoa_nyud_v2} achieve strong results. Our MQTransformers achieve consistent gains for different backbones for each task.
In particular, we observe 2\%-3\% gains on depth prediction and semantic segmentation, 1\%-2\% gains on normal prediction and boundary prediction within an extra 3\% GFlops increase. Moreover, adopting the same backbone and same pre-training, our method consistently outperforms recent work ATRC~\cite{atrc_2021} on HRNet18 with fewer parameters and GFlops. ATRC distills on the largest scale and task interactions do not appear to be sufficient. This implies that our multiple task-relevant query design can adequately capture more dependencies between multiple tasks.
Adopting the Swin-S backbone, our method achieves better results for semantic segmentation and comparable results on normals and boundary prediction. However, the depth prediction on NYUD-v2 is not perfect using Swin, we argue that this is because of the limited dataset size. We believe adding more depth data may lead to better results where similar findings are in previous works~\cite{ViT2021,Ranftl2021,Ranftl2020}.
Our MQTransformer utilizes Swin-L as its backbone, which dramatically enhances MTL performance and surpasses the previous best-performing method (\textit{i.e.,} InvPT~\cite{InvPT_2022}) by +3.08 (mIoU) on SemSeg task.
In addition, our method also achieves the drop in multi-task performance $\Delta_m$. 
As shown in Tab.~\ref{tab:stoa_nyud_v2}, it is clear that MQTransformer outperforms other efficient MTL of dense prediction frameworks dramatically.
For example, MQTransformer using Swin-L achieves +1.28 higher mIoU (54.84 v.s. 53.56) than the competitive InvPT~\cite{InvPT_2022} using ViT-L as a backbone on NYUD-v2 with the fewer GFLOPs (365.25 (ours) v.s. 555.57 (InvPT)) and parameters ((204.3 (ours) v.s. 402.1 (InvPT))) in Tab.~\ref{tab:stoa_nyud_v2}.

\noindent
\textbf{Results on PASCAL-Context dataset.} As shown in Tab.~\ref{tab:stoa_pascal}, we also carry out experiments on PASCAL-Context dataset.
We use four mainstream architectures as the backbone to evaluate our MQTransformer, including HRNet18, ViTAREv2-S, Swin-T and Swin-S.
Again, compared with the strong baseline, our method achieves consistent gains over five different tasks. Moreover, our MQTransformer achieves state-of-the-art results and outperforms previous works by a significant margin.
It can be shown by Tab.~\ref{tab:stoa_pascal} that the proposed method achieves an improved prediction accuracy for different tasks on different backbones.
In addition, we also visualize the feature map in Fig.~\ref{fig:vis_compare}, which shows that the predictions of other methods are coarse.
In contrast, our method's predictions are more fine-grained and have more textures especially on SemSeg and Human Parts tasks.
As shown in Tab.~\ref{tab:stoa_pascal}, Fig.~\ref{fig:vis_compare} and Fig.~\ref{fig:app_pascal}, these observations demonstrate that the proposed MQTransformer grafts the merit of multiple task-relevant queries for capturing highly informative information.

\subsection{Ablation studies and Analysis }
\label{sec:exp_ablation}

\noindent
\textbf{Settings.} For ablations, all models are trained using Swin-S as a backbone with batch size 8 on NYUD-v2 dataset and iteration 40k unless a specific statement is. For visual analysis, we adopt a trained model with Swin-S.

\noindent
\textbf{Network Components.} 
To verify the effectiveness of each key designed modules, we gradually extend the Swin-S baseline to our MQTransformer.
We first consider removing different modules in the whole model in Tab.~\ref{tab:multi-ablaton} (a).
As shown in Tab~\ref{tab:multi-ablaton} (a) w/o QL and w/o CTQA, it indicates that query learning and cross-task query attention benefit segmentation, depth estimation and boundary detection.
Note that when training a model without both QL and CTQA, this will lead to a significant decrease the task performance.
In fact, the full model achieves better accuracy, demonstrating the importance of each module for the final predictions.
In summary, the proposed components of the MQTransformer are each necessary and collectively increase +1.28 mIoU improvements.

\noindent 
\textbf{Depth of MQTransformer.} In Tab.~\ref{tab:multi-ablaton} (b), as the number of the encoder and decoder depth increases, the accuracy does not seem to improve and even tends to decay. This means that only adding one encoder and decoder is enough. 
However, we believe using more data can achieve better results with the increase of depth, which is observed in ~\cite{ViT2021,Ranftl2021}.
we empirically set depth to 1 by default; thus our model is lightweight and efficient.

\begin{figure*}[!t]
\centering
  \includegraphics[width=0.990\textwidth]{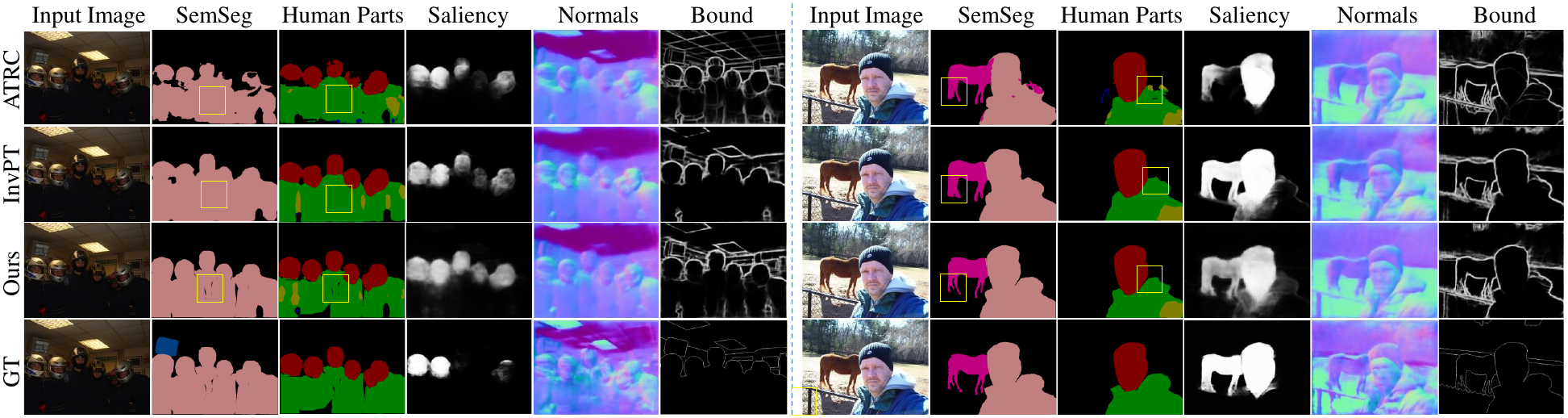}
  \caption{\small Visual comparison results with ATRC~\cite{atrc_2021} model predictions on PASCAL-Context. Our model results in the \textbf{better} prediction on semantic segmentation, human parts segmentation, saliency estimation, surface normal prediction, and boundary detection.
  Please note the human pictures detected in the first three columns.
  } 
  \label{fig:vis_compare}
\end{figure*}

\begin{table}[!t]
  \centering
  \caption{Ablation on shared and non-shared encoder-decoder using different backbones on NYUD-v2.}
  \label{tab:ablation_share}
  \resizebox{.48\textwidth}{!}{
  \begin{tabular}{llllllllll}
  \hline
  \multirow{2}*{Model}  & \multirow{2}*{Backbone}    & {SemSeg}   &Depth & Normals & Bound\\
      &  &(mIoU)$\uparrow$   &(rmse)$\downarrow$   & (mErr)$\downarrow$  &(odsF)$\uparrow$\\
  \hline
  non-shared  &HR18   &39.05   &0.5985   &20.41  &76.50\\
  non-shared  &Swin-S    &48.63   &0.5873   &20.89  &77.00\\
  \hdashline
  shared (Ours) & HR18          &40.47   &0.5965   &20.34   &76.60   \\
  shared (Ours) & Swin-S           &49.18   &0.5785   &20.80    &77.00 \\
  \hline
  \end{tabular}}
\end{table}

\noindent
\textbf{Varying Number of Query $N$.} In Tab.~\ref{tab:multi-ablaton} (c), we change the $N$ within $N\in \{8,32,64,128,156,256\}$. Compared to $N$=8, $N$=64 results in a significant improvement of 2.71\% in segmentation accuracy. However, this improvement saturates when more sections are added to the network at $N$=128.
Notably, larger $N$ is not always better. When $N$=256, it makes an explicit decrease in accuracy for each task. Considering all the metrics, we chose $N$=64 by default.

\noindent
\textbf{Influences of Different Backbones.}
Tab.~\ref{tab:multi-ablaton} (d) presents the effect of the SemSeg, Depth, Normals, and Bound values when adopting the individual backbones. As shown in that table, swin backbones have a high SemSeg value. Notably, our model applies to both CNN~\cite{HRnet_19} and vision Transformer~\cite{swin} backbones and improves accuracy on multiple tasks, indicating the generation ability of our approach. For these different backbones, our method improves the results of different tasks consistently.

\begin{table}[!t]
  \caption{ Ablation on a different number of scales using Swin-S backbone on NYUD-v2.}
  \label{tab:ablation_scales}
  \centering
\begin{tabular}{llllllllll}
\hline
 \multirow{2}*{Model}     & {SemSeg}   &Depth & Normals & Bound\\
      & (mIoU)$\uparrow$   &(rmse)$\downarrow$   & (mErr)$\downarrow$  &(odsF)$\uparrow$\\
\noalign{\smallskip}
\hline
\noalign{\smallskip}
Single-Scale &48.92  &0.5839  &20.88  &75.7\\
Two-Scale    &49.18  &0.5785  &20.80   &77.0\\
Four-Scale   &49.38  &0.6006  &20.99  &76.9\\
\hline
\end{tabular}
\end{table}

\begin{figure}[!t]
\centering
  \includegraphics[width=0.5\textwidth]{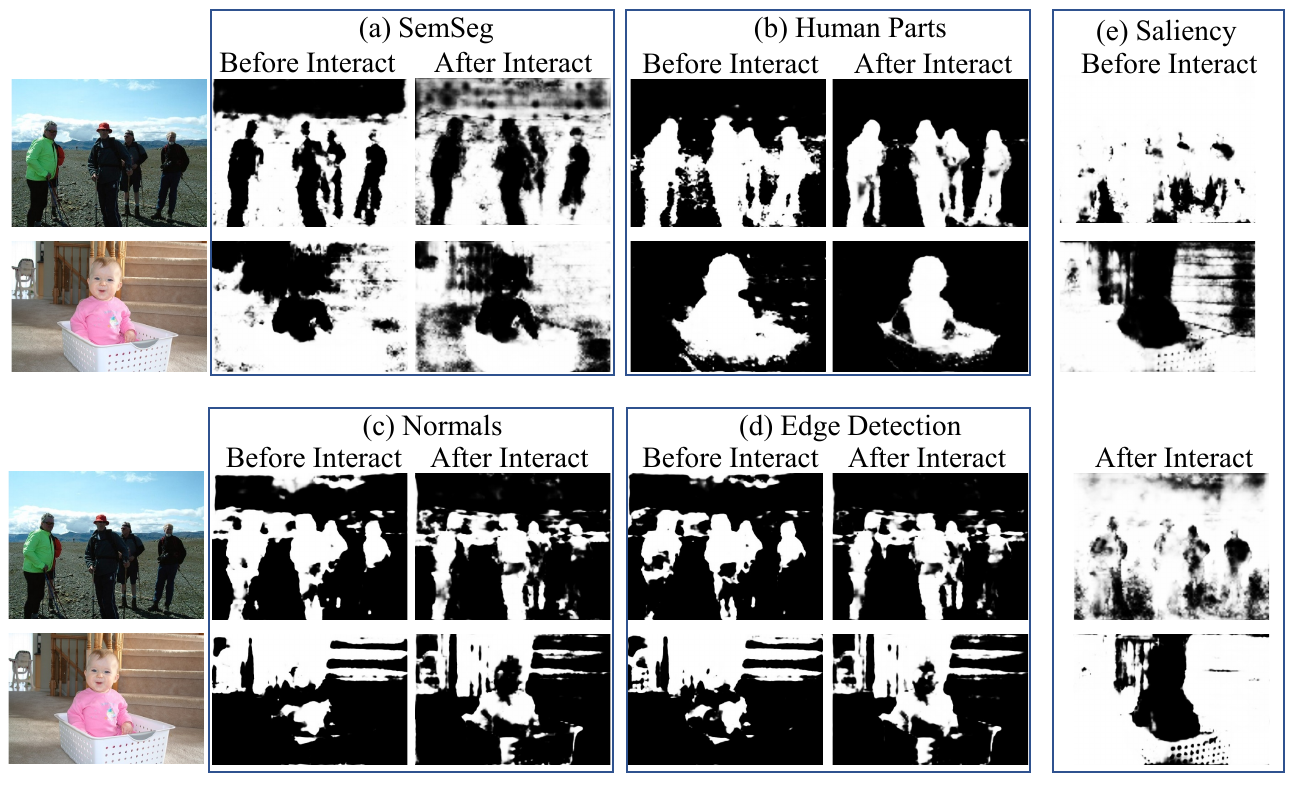}
  \caption{\small Visualization activation maps on PASCAL-Context dataset.
  Qualitative comparison of changes resulting from interactions using cross-task query attention module.
  } 
  \label{fig:vis_atten_query_map}
\end{figure}

\noindent
\textbf{Effect of Shared Encoder and Decoder.} 
We use two mainstream networks as the backbone to evaluate our MQTransformer using shared encoder and decoder, including HRNet18 and Swin-S. 
In Tab.~\ref{tab:ablation_share}, we explore the shared encoder and decoder design where we find that using \textit{a shared encoder and decoder} in MQTransformer leads to better results in different settings with less parameter increase. This verifies our motivation and key design in Sec.~\ref{sec:method}.

\noindent
\textbf{Effect of Scale Number in Shared Feature.}
To generate different scale shared features, we use shared feature $X \in \mathbb{R}^{L\times C}$ (in Fig.~\ref{fig_encoder_decoder}) to perform a sample downsampling operation and obtain $X_2 \in \mathbb{R}^{\frac{L}{4}\times C}$.
In Tab.~\ref{tab:ablation_scales}, we find that two scales are good enough. Compared with a single scale, our design obtains significant gains over four tasks. Adding more scales does not bring extra gains. Thus we set the scale number to 2 by default.

\noindent
\textbf{Visualization on Learned Query Activation Maps.} 
In Fig.~\ref{fig:vis_atten_query_map}, we visualize the attention map for each query in each task. We randomly choose one query for visualization. As shown in that figure, after the cross-task query interaction, we found more structures and fine-grained results for each task, proving the benefits of our query-based cross-task attention design.
The variations in query activation maps demonstrate the effectiveness of the proposed cross-task query attention module.
In addition, the results show that a task-relevant query-based Transformer is beneficial to multi-task learning of dense prediction tasks.

\begin{figure*}[!t]
\centering
  \includegraphics[width=0.98\textwidth]{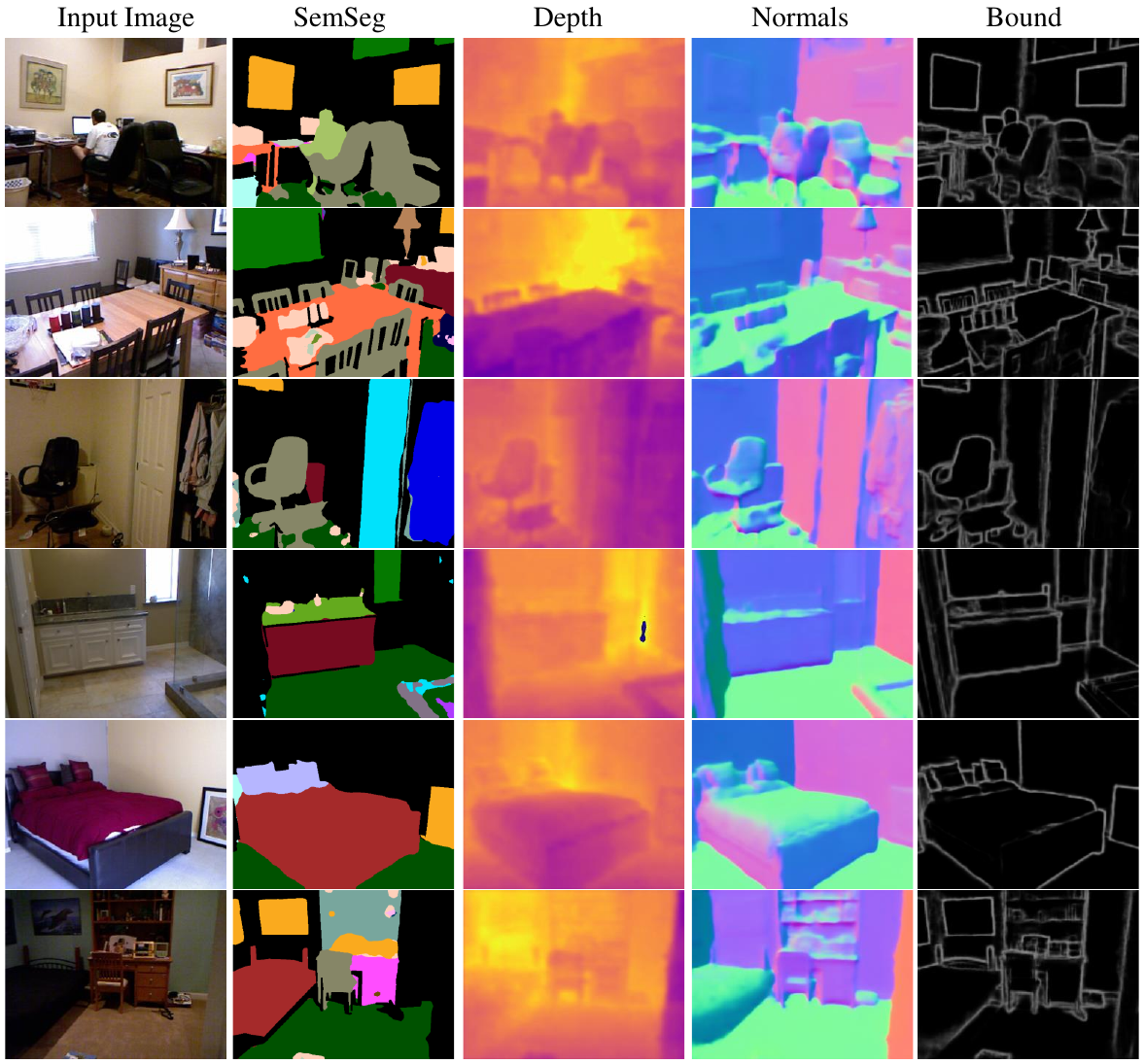}
  \caption{\small Visualization results on NYUD-v2 dataset.
  Our model can result in the correct prediction on semantic segmentation, depth estimation, surface normal prediction, and boundary detection tasks.} 
  \label{fig:vis_nyud}
\end{figure*}

\noindent
\textbf{Visualization Comparison with Previous Methods.} We also present several visual comparisons with recent works on PASCAL-Context dataset in Fig.~\ref{fig:vis_compare}. Compared with recent work~\cite{atrc_2021,InvPT_2022}, our method has better visual results for all five tasks. The advantage of our method is that the semantic segmentation and human parts segmentation tasks show more reliable segmentation.
Fig.~\ref{fig:vis_compare} shows some exemplars of prediction results and it suggests that our model makes effective use of a query-based transformer.
Our model generates dense predictions with better semantic segmentation details, as marked in the yellow square box. 
We can observe from Fig.~\ref{fig:vis_compare} that our method’s predictions are more fine-grained and have more local edges and textures especially on SemSeg and Human Parts segmentation tasks.
In addition, we can see that our MQTransformer can successfully segment the details of a single person.
Fig.~\ref{fig:vis_compare} and Tab.~\ref{tab:stoa_pascal} indicate that our MQTransformer could work finely.

\begin{figure*}[!t]
\centering
  \includegraphics[width=0.95 \textwidth]{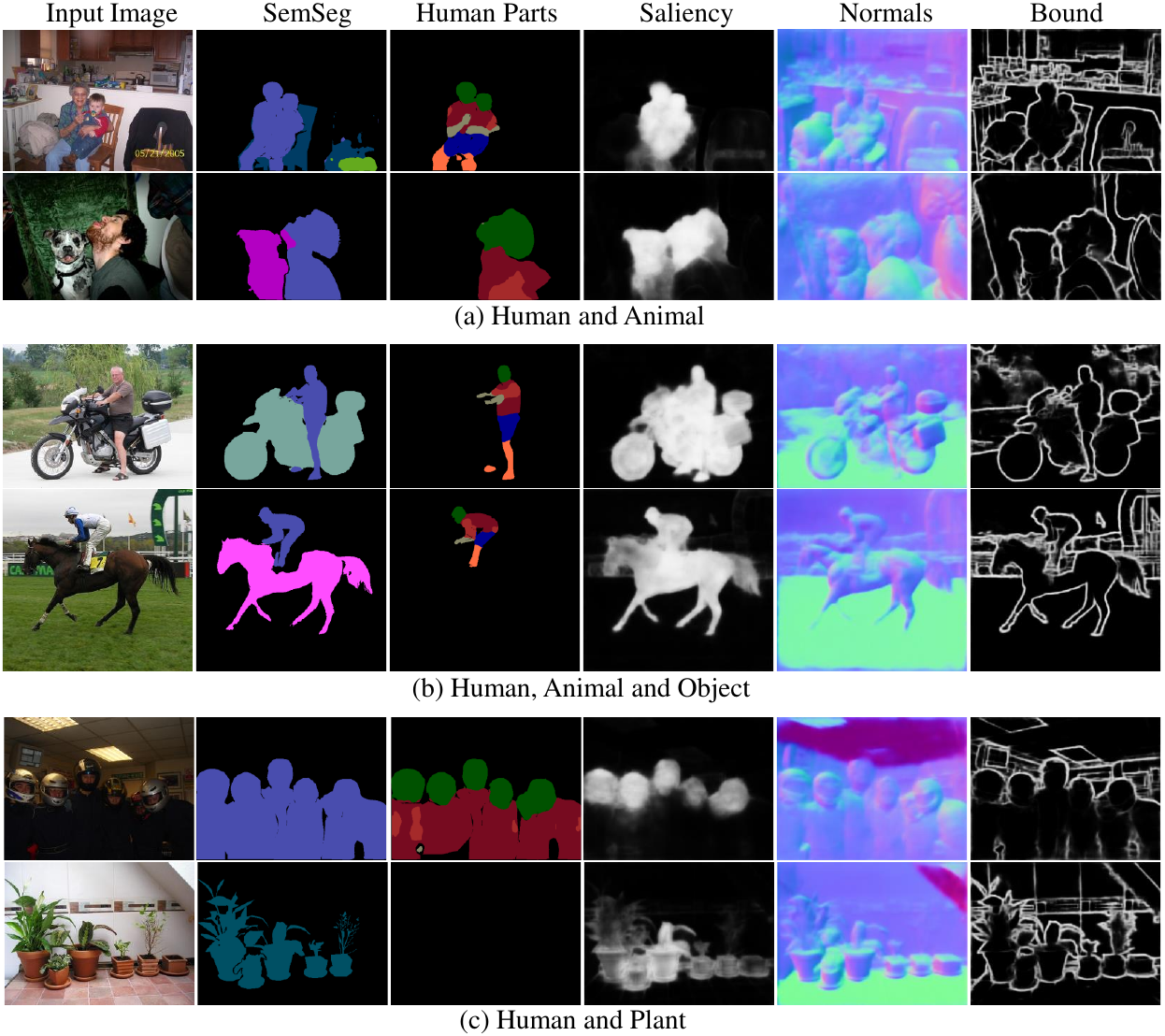}
  \caption{\small Qualitative results on PASCAL-Context dataset.
  We group the visualization results into three groups for comparison. (a) Human and animal; (b) Human, animal and object; (c) Human and plant.
  Our model is able to find the correct image feature corresponding to different tasks and eventually get the correct prediction on semantic segmentation (SemSeg), human parts segmentation (Human Parts), saliency estimation (Saliency), surface normals prediction (Normals), and boundary detection (Bound) tasks.
  } 
  \label{fig:app_pascal}
\end{figure*}

\noindent
\textbf{Visualization Results on NYUD-v2.}
For a more vivid understanding of our model, Fig.~\ref{fig:vis_nyud} shows images of the qualitative results of our MQTransformer with a Swin-S backbone on the NYUD-v2 dataset. Our method also achieves strong visualization results.

\noindent
\textbf{Visualization Results on PASCAL-Context.}
As shown in Fig.~\ref{fig:app_pascal}, we group the images into three groups for a stronger visual contrast. Fig.~\ref{fig:app_pascal} (a) shows two sets of images of a woman holding a child and a man holding a dog. On the human parts segmentation task, our model segments the two people in the first set of images, while the second set of images only segments the human part, ignoring the dog.
Fig.~\ref{fig:app_pascal} (b) also verifies the accuracy of the human parts segmentation task.
In the last row of images (see Fig.~\ref{fig:app_pascal} (c)), it is worth noting that the third image is all black since this one is all plants with no human parts. 
Moreover, in other tasks, qualitative results also demonstrate good visual performance.

\noindent
\textbf{Limitation and Future Work.}
We demonstrate the advantages of our proposed MQTransformer. However, one limitation of our method is that our task-relevant query features are randomly initialized.
As shown in Tab.~\ref{tab:multi-ablaton} (c), we show the effect of the $N$ of the task-relevant query feature ($P\in \mathbb{R}^{N \times C}$) on the model.
In the future, we will first generate task-relevant query features using image information.
Thus, more task-relevant and fine details can be obtained.
We will be exploring the task-relevant query features and the potential of the Transformer architecture in MTDP.

\noindent
\textbf{Boarder Impact.} Our method explores multi-task dense prediction with a novel multi-query transformer architecture. It is a new encoder-decoder baseline for this task and may inspire the new design of the multi-task learning framework.

\section{Conclusion}
\label{sec:conclusion}

In this paper, we propose a new vision transformer architecture for MTDP. 
We design novel multiple task-relevant queries to extract task-specific features from different tasks. Then we perform task association via cross-task query attention which avoids huge pixel-level computation and cost that are used in previous works.
Then a shared encoder and decoder network is adopted to exchange information between queries and corresponding task-specific features. 
Extensive experiments show that our model can achieve significant improvements on different metrics with various strong baselines.
Moreover, we show the effectiveness of our method on two multi-task learning datasets, using CNN \& Transformer backbones and efficient architectures across multiple vision tasks.
We hope our method can be a new simple yet effective transformer baseline for MTDP.

\noindent
\section*{Data availability} 
The datasets generated during and/or analysed during the current study are available in the NYUD-v2 and PASCAL-Context repositories, ~\url{https://cs.nyu.edu/~silberman/datasets/nyu_depth_v2.html} and \url{https://www.cs.stanford.edu/~roozbeh/pascal-context/}

\bibliographystyle{IEEEtran}
\bibliography{egbib}


\vfill

\end{document}